\documentclass[conference]{IEEEtran}
\IEEEoverridecommandlockouts
\usepackage{cite}
\usepackage{amsmath,amssymb,amsfonts}
\usepackage{algorithmic}
\usepackage{graphicx}
\usepackage{textcomp}
\usepackage{xcolor}
\usepackage{subcaption}
\usepackage{multirow}
\usepackage{tabularx}
\usepackage{hhline}
\usepackage{booktabs}
\usepackage{siunitx}
\usepackage{arydshln}    
\let\oldcite\cite
\renewcommand{\cite}[1]{\textcolor{blue}{\oldcite{#1}}}
\usepackage{hyperref}
\hypersetup{
    colorlinks=true,
    citecolor=blue,
    linkcolor=blue,
    urlcolor=blue
}

\def\BibTeX{{\rm B\kern-.05em{\sc i\kern-.025em b}\kern-.08em
    T\kern-.1667em\lower.7ex\hbox{E}\kern-.125emX}}
\begin{document}

\title{Hypergraph Contrastive Sensor Fusion for
Multimodal Fault Diagnosis in Induction Motors}
\author{Usman Ali, Ali Zia, Waqas Ali, Umer Ramzan, Abdul Rehman, Muhammad Tayyab Chaudhry, and \\
Wei Xiang \IEEEmembership{Senior Member, IEEE}}

\maketitle

\begin{abstract}

Reliable induction motor (IM) fault diagnosis is vital for industrial safety and operational continuity, mitigating costly unplanned downtime. Conventional approaches often struggle to capture complex multimodal signal relationships, are constrained to unimodal data or single fault types, and exhibit performance degradation under noisy or cross-domain conditions. This paper proposes the Multimodal Hypergraph Contrastive Attention Network (MM-HCAN), a unified framework for robust fault diagnosis. To the best of our knowledge, MM-HCAN is the first to integrate contrastive learning within a hypergraph topology specifically designed for multimodal sensor fusion, enabling the joint modelling of intra- and inter-modal dependencies and enhancing generalisation beyond Euclidean embedding spaces. 
The model facilitates simultaneous diagnosis of bearing, stator, and rotor faults, addressing the engineering need for consolidated diagnostic capabilities. Evaluated on three real-world benchmarks, MM-HCAN achieves up to 99.82\% accuracy with strong cross-domain generalisation and resilience to noise, demonstrating its suitability for real-world deployment. An ablation study validates the contribution of each component. MM-HCAN provides a scalable and robust solution for comprehensive multi-fault diagnosis, supporting predictive maintenance and extended asset longevity in industrial environments.
\end{abstract}


\section{Introduction}
\label{sec:introduction}
\IEEEPARstart{I}{nduction} motors (IMs) are essential to modern industrial systems, supporting sectors like manufacturing, energy, and transportation. However, faults in IMs can cause downtime, high maintenance costs, and substantial economic losses.
As a result, fault diagnosis in IMs has become a focal point of research, with recent studies highlighting its importance in enhancing operational resilience and minimising financial impacts. IMs faults are broadly classified as either electrical, with stator faults comprising 28-36\%, or mechanical, encompassing bearing (42-55\%) and rotor (8-10\%) failures \cite{b1}.
Detecting these faults requires a systematic analysis of motor signals, such as current, voltage, and vibration. The accuracy of fault classification depends heavily on selecting appropriate signal types and employing advanced data acquisition techniques that provide actionable insights into the motor's condition. Among these techniques, current monitoring and vibration signal analysis have gained prominence due to their non-intrusive nature, sensitivity, and reliability \cite{b2}.
Traditional fault diagnosis techniques, time/frequency domain analysis \cite{ali3} and wavelet transforms, offer simplicity but struggle with complex fault patterns \cite{up1,up2,up3,up4}.

Data-driven methodologies \cite{up6, up7,ali1}, particularly machine learning (ML), have demonstrated significant potential in capturing complex nonlinear relationships within fault data from rotating machinery. In the domain of bearing fault diagnosis, for instance, ensemble learning (EL) strategies have been explored to refine classification accuracy. Illustratively, the authors in \cite{b108} adopted an EL approach, merging random forest and extreme gradient boosting algorithms. This method has been notably developed and validated on a limited-scale multi-class dataset. Similarly, an EL architecture employing the archimedes optimisation algorithm (ArchOA) with gradient boosting decision trees (GBDT) demonstrated 97.50\% accuracy for compound bearing fault detection \cite{br2}. However, its training on a mere 250 samples raises concerns about potential overfitting. Other ML techniques, such as a layered feature extraction methodology combining discrete wavelet transform (DWT) with binary signatures and nearest component analysis (classified by SVM and KNN) \cite{b18,ali2}, and a DWT-genetic algorithm framework (GaBoT) \cite{br3}, have reported high accuracies (99.8\% and 99.18\%, respectively). Nevertheless, these approaches also face limitations, including evaluation on fewer test sets and high computational complexity \cite{br3}.
 For stator and rotor fault diagnostics, similar limitations often emerge. An ML-based approach for stator faults, utilising AdaBoost on fused time-domain features from current and vibration signals. This work is hindered by a limited dataset and its exclusive reliance on time-domain features, suggesting that future enhancements could benefit from incorporating frequency-domain information \cite{st1}. In rotor fault detection, an optimised Stockwell transform achieved 97.41\% accuracy for two broken rotor bars (BRB) \cite{rc1}. Despite the availability of both current and vibration signals (including for 4-BRB conditions). This analysis has been confined only to vibration data, indicating that fusing both modalities could improve diagnostic performance and generalisation. Likewise, an SVM-based classifier using FFT-enhanced current signal features for BRB diagnosis attained 95.80\% accuracy \cite{rc3}. However, integrating vibration signal data through multimodal feature fusion presents a clear direction for further performance enhancement. \textit{While individual ML models show promise, their efficacy is often limited by dataset size or a unimodal analytical approach, thereby pointing towards the critical need for methods that can effectively leverage multiple data sources and generalise well, even from potentially limited data.}

Concurrently, deep learning (DL) models, including convolutional neural networks (CNNs), long short-term memory (LSTM) networks, and autoencoders, have gained prominence for their exceptional performance in handling high-dimensional fault data. Despite their success, these models primarily rely on sequential interactions, which inherently limit their ability to capture higher-order dependencies among fault features \cite{b7q}. Moreover, conventional DL approaches typically analyse raw signals (e.g., current, vibration) and spectral images in isolation, thereby failing to exploit the complementary insights that could be derived from integrating these modalities. Zhang et al. \cite{b100} proposed a deep CNN framework for bearing faults detection that achieved improved accuracy even in noisy and varying workload conditions through advanced training and refined architectures. Nonetheless, the data augmentation techniques employed carry a risk of introducing additional noise, potentially affecting model robustness. For stator fault detection \cite{S1}, a 2D CNN analysing fundamental frequency phasor magnitudes and third harmonic components from stator current signals offers robust detection on inter-turn short circuits, acknowledging that limited training data availability might impact model generalizability.  
Hybrid models, such as a DWT-integrated CNN with LSTM-governed weight updates \cite{st2}  for stator fault diagnosis (98.20\% accuracy) , and a Hilbert transform with a dual-branch fusion residual CNN (DBF-CNN) for BRB detection (99\% accuracy) \cite{rc4}, predominantly have relied on single-modality current signals. Similarly, a MobileNetV2 architecture using STFT-based spectrograms from vibration signals for BRB classification (97.78\% accuracy) \cite{rc2} also focuses on a single data source. In \cite{wpedl}, the authors developed a weighted probability ensemble DL technique for multi-class, cross-domain fault generalisation on high-dimensional data. One consideration for this model is the relatively high computational time needed for decision evaluation. \textit{Consequently, these DL studies consistently point to limitations stemming from data constraints, potential noise introduction via augmentation, and the underutilization of multimodal data fusion, a gap that could significantly enhance diagnostic robustness and reliability across diverse operating conditions.}

Graph Neural Networks (GNNs) have emerged in fault classification, with applications ranging from STFT-based label sampling \cite{gnn2} to few-shot learning frameworks \cite{gnn1}. However, GNNs are restricted to pairwise feature interactions, limiting performance on complex multimodal data. Hypergraph Neural Networks (HGNNs) address this by modelling higher-order relationships across modalities \cite{az3}, yet are underexplored in industrial diagnosis and often lack feature discrimination mechanisms.

Concurrently, contrastive learning (CL) has gained traction as a powerful technique for enhancing feature representation in industrial fault diagnosis~\cite{az2}. CL can significantly improve model generalisation by maximising inter-class separation while preserving intra-class consistency. However, prior CL approaches are predominantly applied to Euclidean space embeddings.

Our proposed method, MM-HCAN, bridges these gaps by uniquely integrating HGNN-based contrastive learning with a multi-head attention mechanism, specifically tailored for multimodal industrial datasets. MM-HCAN is characterised by its construction of separate intra-modality and cross-modality hypergraphs. This approach explicitly models both dependencies within a single data type and relationships between different modalities, facilitating deeper feature fusion and more precise fault localisation. This contrasts with many existing hypergraph methods that may not fully exploit such rich, multi-faceted dependencies crucial for robust classification. Furthermore, the incorporation of multi-head attention refines feature discrimination, further bolstering MM-HCAN's resilience against noisy industrial signals.
\textit{To the best of our knowledge, MM-HCAN is the first approach to apply hypergraph contrastive learning with multi-head attention in an industrial fault diagnosis setting.}

The proposed architecture processes raw signals and STFT images from rotating machinery to diagnose faults using a dual-pathway approach. The raw signal is analysed temporally through 1D CNNs and an LSTM, while the STFT image is processed spectrally using the ResNet module. Both pathways extract 512-dimensional feature vectors representing temporal and spectral information. A hypergraph-based framework integrates these features by treating each feature dimension as a node and connecting them via hyperedges. Hyperedges are KNN formed using similarity measures like cosine similarity, with separate hypergraphs for intra-modality (temporal or spectral) and cross-modality (between temporal and spectral) interactions. An HGNN updates the embeddings to capture higher-order relationships. 

To further enhance feature discriminability, a contrastive learning-based triplet loss function is employed, ensuring that similar samples are positioned closer together in the embedding space while dissimilar ones are pushed apart. Finally, a multi-head attention mechanism fuses the temporal, spectral, and cross-modality embeddings into a unified representation, which is subsequently fed into a classification network to predict fault categories. This approach facilitates robust fault diagnosis across diverse operational conditions, making MM-HCAN highly effective in real-world industrial applications.
The following key \emph{contributions} of this work are mentioned below:

\begin{itemize}
  \item We propose MM-HCAN, a unified framework for the simultaneous classification of bearing, stator, and rotor faults using multimodal signal fusion, eliminating the need for separate fault-specific models.
  
  \item We introduce a novel hypergraph-based contrastive learning approach that models both intra- and cross-modality relationships, enhancing discriminative learning across temporal and spectral domains.
  
  \item We integrate a multi-head attention mechanism to refine feature selection and improve interpretability, further boosting classification robustness under noisy and cross-domain conditions.
  
  \item We conduct extensive experiments on benchmark datasets, including detailed ablation studies that validate each architectural component. These demonstrate MM-HCAN's superior accuracy, robust generalisation, and noise resilience over state-of-the-art models.

\end{itemize}

The remainder of this paper is organised as follows. Section~\ref{sec:pm} presents the MM-HCAN architecture, including preprocessing, feature extraction, and hypergraph-based learning. Section~\ref{sec:exsetup} describes the experimental setup, covering datasets, training configuration, and STFT parameters. Section~\ref{sec:res} evaluates MM-HCAN performance through individual and cross-domain classification, robustness tests, benchmarking, ablation studies, and efficiency analysis. Section~\ref{sec:con} concludes the paper. Additional architectural analysis, extended results, and discussions are provided in the \textit{Supplementary Material}.

\section{Proposed Methodology}\label{sec:pm}
\begin{figure}[t]
    \centering
    \includegraphics[width=0.45\textwidth]{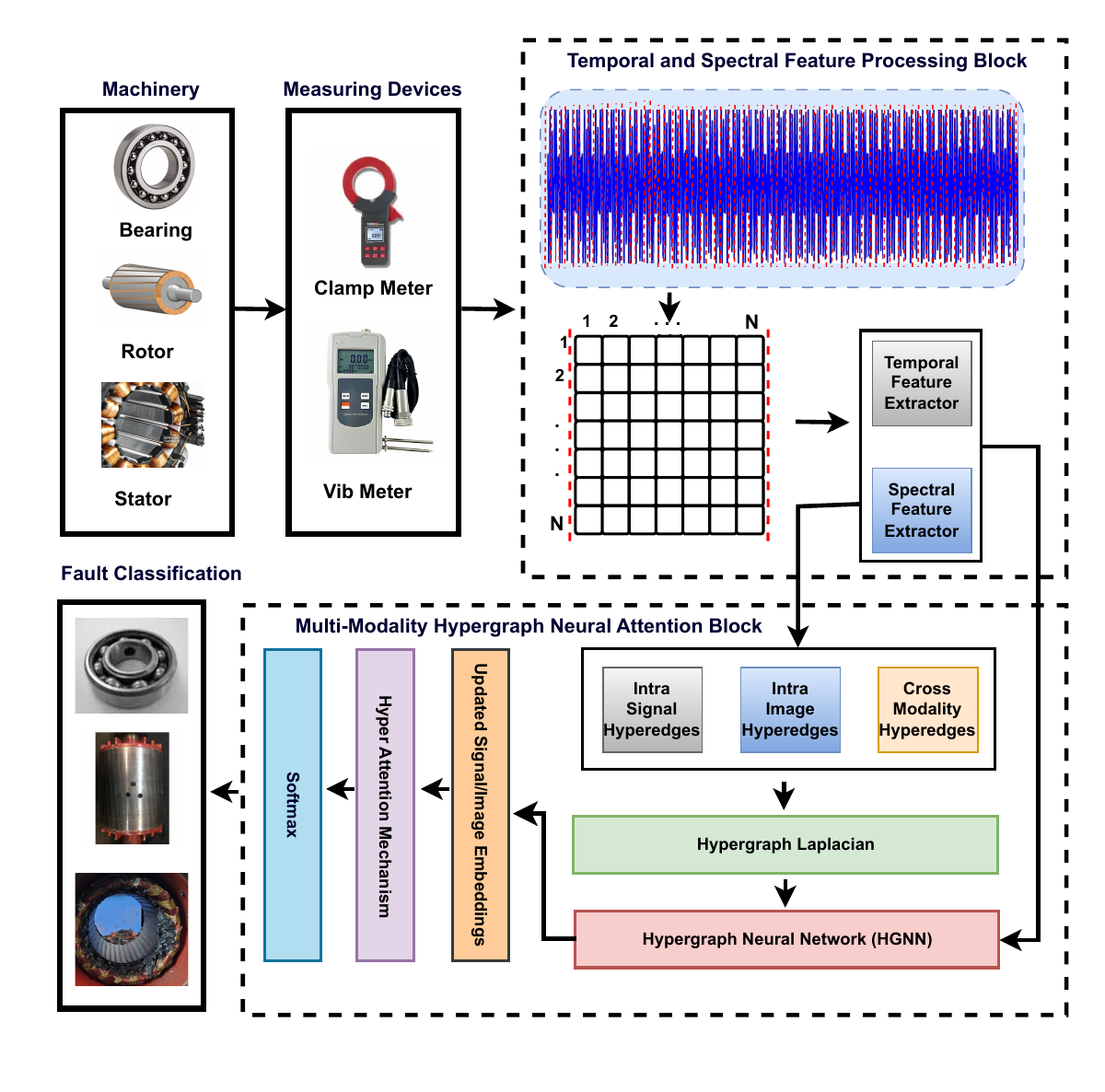} 
    \caption{MM-HCAN architecture: temporal features from raw signals (1D CNN-LSTM) and spectral features from STFT images (ResNet-18) are fused via hypergraph contrastive learning and multi-head attention for classification.}
    \label{fig1}
\end{figure}

Figure \ref{fig1} illustrates the comprehensive framework of our proposed methodology. The process begins with the acquisition of raw signals from various IM components, which are measured using clamp sensors and vibration meters. These signals are then segmented into fixed time intervals, and each segmented signal is processed through two distinct feature extraction modules: temporal feature extraction (raw signal) and spectral feature extraction (STFT image). The extracted features are the foundation for constructing two types of hypergraphs: intra-modality hypergraphs and cross-modality hypergraphs. Intra-modality hypergraphs capture the relationships within individual modalities (e.g., temporal or spectral), while cross-modality hypergraphs model the interactions between different modalities.

\begin{figure*}[t]
    \centering
    \includegraphics[width=0.95\textwidth]{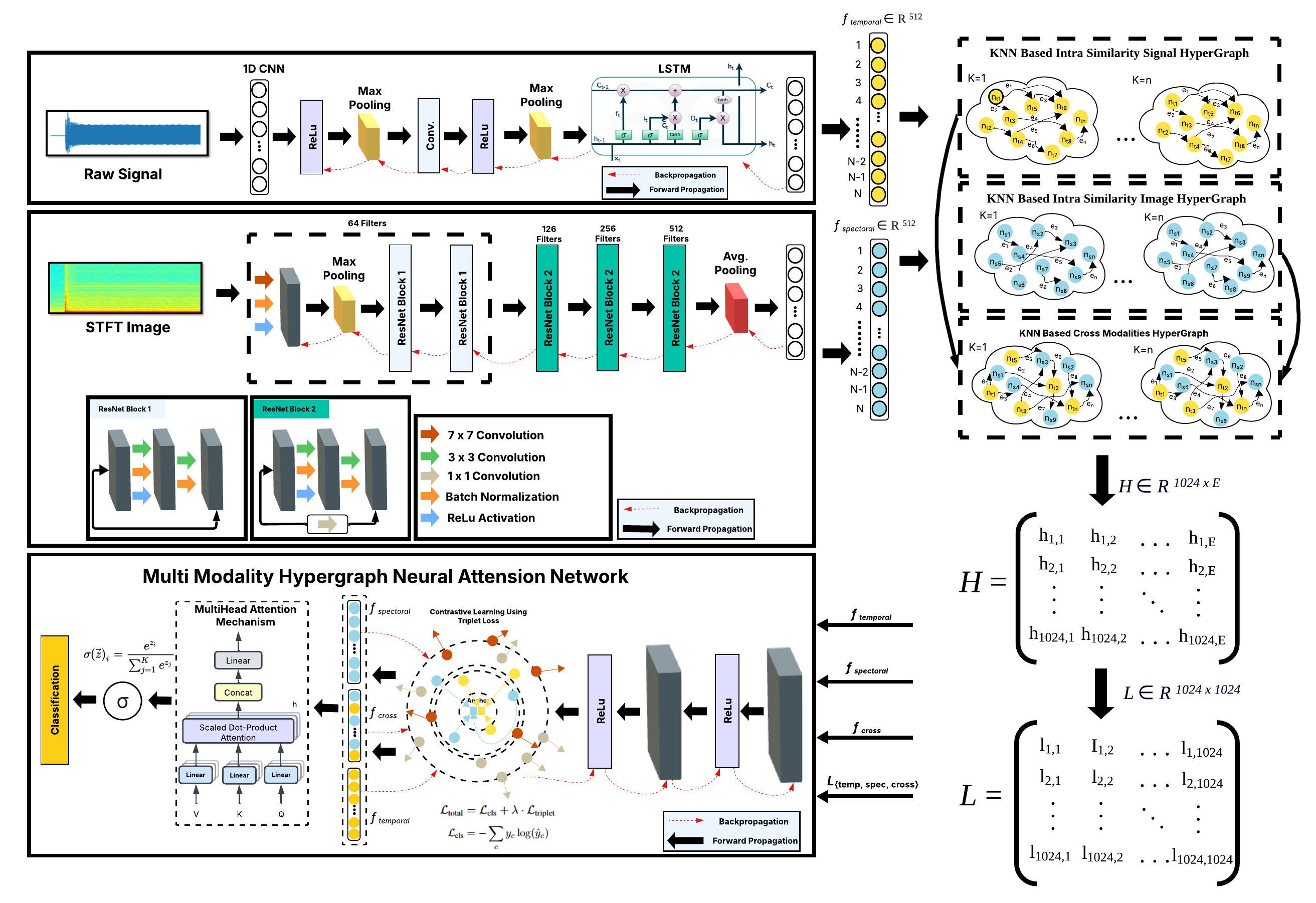} 
    \caption{Hypergraph construction: intra- and cross-modality hyperedges are formed via KNN to model high-order relationships among feature vectors across modalities. }
    \label{fig2}
\end{figure*}

 To enhance the representational power, the temporal and spectral embeddings are concatenated, forming a unified feature representation. These features, along with the hypergraph Laplacian matrices, are fed into a contrastive-based learning two-layer HGNN. This network generates updated embeddings that encapsulate both local and global structural information from the input data. To further refine the feature representations, a multi-head attention mechanism is applied. This mechanism enables the model to focus on the most discriminative features across modalities by computing attention weights dynamically. The resulting feature vectors are then passed through a softmax classifier to produce the final classification output. The subsequent sections in Figure \ref{fig2} provide an in-depth analysis of the individual modules, including signal processing, feature extraction, construction of hypergraphs, the design of the HGNN layers, and the implementation of the attention mechanism.

\subsection{Preprocessing Block}
The dataset comprises vibration and current signals collected from rotating machinery under various fault conditions, such as bearing, stator, and rotor faults. To ensure a consistent representation, continuous time-series signals are divided into non-overlapping sequences of fixed length $T$, where the segmentation process for a given signal $S = \{S_1, S_2, \dots, S_n\}$, with each segment $S_i$ belonging to $\mathbb{R}^T$ and $n$ representing the total number of segments. This ensures balanced class representation by maintaining an equal number of signal segments and spectrograms across different fault categories. Each segment $S_i$ is analysed through two parallel pathways: temporal analysis using raw signals and spectral analysis via STFT-generated spectrograms. To mitigate amplitude variations, each segment $S_i$ is standardized to zero mean and unit variance using the normalization $S_{\text{norm},i} = \frac{S_i - \mu_i}{\sigma_i}$, where $\mu_i$ and $\sigma_i$ denote the segment-wise mean and standard deviation, respectively, ensuring that the data is appropriately scaled for subsequent feature extraction and model training. Each \( S_i \) is converted to a time-frequency representation \( X_i(t, f) \) using the STFT equation:
\begin{equation}
X_i(t, f) = \sum_{n=-\infty}^{\infty} S_i(n) \, w(n - t) \, e^{-j 2 \pi f n},
\end{equation}
where \( w(n - t) \) is a window function localising the signal in time. To accentuate low-magnitude frequency components, a log transform is applied:
\begin{equation}
X'_i(t, f) = \log\left(1 + |X_i(t, f)|\right).
\end{equation}
The spectrograms \( X'_i(t, f) \) are resized to uniform dimensions and normalised to \([0, 1]\) via min-max scaling to ensure compatibility with DL architectures.

\subsection{Temporal and Spectral Feature Block} \label{T_S_F}
The feature extraction framework transforms raw temporal signals into a unified 512-dimensional representation using a dual-stream architecture. For temporal processing, normalised input segments $ S_{\text{norm},i} \in \mathbb{R}^T $ are processed through two sequential 1D CNN layers followed by an LSTM network. The first CNN layer uses 64 filters with a kernel size 7 and stride 1, while the second layer applies 128 filters with kernel size 5 and stride 1. The generic feature map at each layer is given by $ C_i = \text{ReLU}(\text{Conv1D}(\text{MaxPool}(C_{i-1}); W_i, b_i)) $, where $ C_0 = S_{\text{norm},i} $, and $ W_i, b_i $ are learnable parameters. A max pooling operation (pool size 2, stride 2) reduces spatial dimensions. These features are then passed to an LSTM, producing hidden states $ h_t = \text{LSTM}(C_2; \theta_{\text{LSTM}}) $, where $ h_t $ represents the hidden state at timestep $ t $.

In parallel, spectral processing converts log-scaled spectrograms $ X'_i(t,f) $ into 512-dimensional vectors using a ResNet-18 architecture. ResNet-18 is chosen for spectral feature extraction due to its balance between feature expressiveness and computational efficiency, whereas deeper models (e.g., ResNet-50) increase complexity without significant accuracy gains. For temporal analysis, 1D CNNs are preferred over LSTMs, as they efficiently capture local patterns while avoiding high training complexity and vanishing gradient issues in long time-series data. The inclusion of both 1D CNN and LSTM enables complementary extraction of localised and temporal-sequential patterns, while ResNet-18 efficiently captures spectral discriminative features. This diversity enhances MM-HCAN’s ability to handle heterogeneous signal dynamics.

For time-frequency representation, STFT is employed instead of wavelet transforms, as it provides fixed time-frequency resolution, making it ideal for IM signals where fault patterns exhibit consistent frequency shifts. Wavelet transforms require careful selection of mother wavelets, introducing subjective bias in feature extraction. The combination of 1D CNNs for temporal feature extraction, STFT for time-frequency analysis, and ResNet for spectral representation ensures that MM-HCAN effectively captures both temporal and spectral fault characteristics, leading to superior classification performance.

\subsection{Hypergraph-Based Multi-Modal Fusion}

To integrate temporal ($ f_t \in \mathbb{R}^{512} $) and spectral ($ f_s \in \mathbb{R}^{512} $), and cross ($ f_c \in \mathbb{R}^{1024} $) feature representations, a structured hypergraph-based framework is constructed. Each feature dimension is treated as a node, interconnected through hyperedges defining relationships within and across modalities. KNN dynamically identifies neighbours based on similarity:
\begin{equation}
\text{Sim}(X_i, X_j) = \frac{X_i \cdot X_j}{\|X_i\| \|X_j\|}.
\end{equation}

Hyperedges are formed by connecting nodes to their top-$ K $ nearest neighbours. Separate thresholds govern intra-modality ($ \theta_{\text{intra}} $) and cross-modality ($ \theta_{\text{cross}} $) connections. The hypergraph is represented using an incidence matrix $ H \in \mathbb{R}^{N \times E} $:
\begin{equation}
H[i, j] =
\begin{cases}
1 & \text{if node } i \text{ belongs to hyperedge } j, \\
0 & \text{otherwise}.
\end{cases}
\end{equation}
where \textit{N} is the number of nodes and \textit{E} is the number of edges in the hypergraph.
The $H_t \in \mathbb{R}^{512 \times E_t}$,  $H_s \in \mathbb{R}^{512 \times E_s}$, and $H_c \in \mathbb{R}^{1024 \times E_t}$ are defined for the temporal, spectral and cross modalities, respectively. From these incidence matrices, the corresponding hypergraph laplacians ($L_t \in \mathbb{R}^{512 \times 512}$, $L_s \in \mathbb{R}^{512 \times 512}$, and $L_c \in \mathbb{R}^{1024 \times 1024}$) are computed. For any given hypergraph, its Laplacian $L$ is calculated as:
\begin{equation}
L = I - D_v^{-1/2} H D_e^{-1} H^T D_v^{-1/2},
\end{equation}
where $D_v$ is the diagonal matrix of node degrees (i.e., $D_v[i, i] = \sum_j H[i, j]$) and $D_e$ is the diagonal matrix of hyperedge degrees (i.e., $D_e[j, j] = \sum_i H[i, j]$).

These Laplacians ($L_t, L_s, L_c$) and initial feature vectors ($f_t, f_s, f_c$ ) are subsequently fed into a two-layer HGNN. The HGNN propagates information according to the general rule $X'^{(l)} = \text{ReLU}(L X'^{(l-1)} W^{(l)})$. The temporal and spectral embeddings are updated as:
$f_m^{(l+1)} = \text{ReLU}(L_m \cdot f_m^{(l)} \cdot W_m^{(l)}), where \quad m \in \{t, s\}$, and 
 for cross-modality, the concatenated embeddings $f_c$ are updated using $L_c$:
$f_c^{(l+1)} = \text{ReLU}(L_c \cdot F_c^{(l)} \cdot W_c^{(l)})$.

A triplet loss function is incorporated into the training process to enhance the discriminative capability of the learned multimodal representations. The primary goal of the triplet loss is to organise the embedding space such that features corresponding to the same class are clustered closely (minimising intra-class variance), while features from different classes are pushed further apart (maximising inter-class separability). For each modality $m \in \{t, s, c\}$, triplets of samples are considered, consisting of an anchor sample ($a_m$), a positive sample ($p_m$) belonging to the same class as the anchor, and a negative sample ($n_m$) belonging to a different class.  The triplet loss, denoted as $\mathcal{L}_{\text{triplet}}$, is then formulated to penalise embeddings where the distance between the anchor and the positive sample is not sufficiently smaller than the distance between the anchor and the negative sample. Triplet loss is integrated into the HGNN framework for all embeddings ($ f_t^{(t+1)} $, $ f_s^{(t+1)} $, $ F_c^{(t+1)} $) is expressed as:
\begin{equation}
\mathcal{L}_{\text{triplet}} = \sum_{m \in \{t, s, c\}} \max(0, d(a_m, p_m) - d(a_m, n_m) + \alpha),
\end{equation}
The function $d(x, y) = \|x - y\|$ denotes the euclidean distance between the vector embeddings $x$ and $y$.

The overall training objective of the model, $\mathcal{L}_{\text{total}}$, is a composite loss function that combines the cross entropy classification loss ($\mathcal{L}_{\text{cls}}=- \sum_c y_c \log(\hat{y}_c)$) with triplet loss ($\mathcal{L}_{\text{triplet}}$). This is formulated as:
\begin{equation}
\mathcal{L}_{\text{total}} = \mathcal{L}_{\text{cls}} + \lambda \cdot \mathcal{L}_{\text{triplet}}
\label{eq:total_loss}
\end{equation}
where $\lambda$ is a hyperparameter that balances the contribution of the triplet loss relative to the primary classification task. While MM-HCAN applies supervised triplet loss using class labels, the hypergraph construction itself is self-supervised, relying solely on feature similarity to define higher-order relationships. Future extensions may explore fully self-supervised contrastive objectives to reduce dependence on labelled data and enhance generalisation to novel fault types.

After HGNN processing, a multi-head attention mechanism fuses the embeddings:
\begin{equation}
f'_{\text{fused}} = \sum_{m \in \{t, s, c\}} \alpha_m f'^{(2)}_m, \quad \alpha_m = \text{softmax}(W_m f'^{(2)}_m).
\end{equation}
where $\alpha_m$ is attention weights and $W_m$ represents a learnable weight matrix for modality $m$.
The fused representation is passed through a fully connected network for final classification:
\begin{equation}
\hat{y} = \text{softmax}(W_{\text{dense}} f'_{\text{fused}} + b),
\end{equation}
where  $ W_{\text{dense}} $ is the weight matrix and $ b $ is the bias term.

\section{Experimental Setup}\label{sec:exsetup} 

\subsection{Dataset Description}
Three open-source datasets (i.e., rotor \cite{data1}, stator \cite{data2}, and bearing \cite{data3}) have been utilised in this research work. The details of each category are briefly explained below:
\subsubsection{Rotor Dataset Description}
The rotor dataset contains a 1-horsepower IM operating at voltages of 220V / 380V and discharge currents of 3.02A / 1.75A. It has four poles that operate at a frequency of 60 HZ and has a rotation of 1715 rpm. Experiments include load capacities on 12.5\%, 25\%, 37.5\%, 50\%, 62.5\%, 75\%, 87.5\%, and 100\%. Using AC probes with a capacity of up to 50ARMS with an output voltage of 10 mV/A. Five axial accelerometers are used for mechanical signal evaluation.  They feature a frequency range from 5 to 2000 Hz, a sensitivity of 10 MV/mm/s.
Under each loading condition, signals are sampled simultaneously for up to 18 seconds and repeated ten times. The data contains information about four rotor classes for analysis: healthy and one, two, three, and four BRB faults.

\subsubsection{Stator Dataset Description}
The stator dataset includes vibration and current data from three PMSMs (1.0 kW, 1.5 kW, and 3.0 kW). Each motor exhibits between inter-coil circuit faults and inter-turn circuit faults. These motors run at 3000 RPM under a load that limits the torque to 15\% (1.5 Nm). Vibration data was collected with accelerometer sampling at 25.6 kHz for 120 seconds, while CT sensors recorded current data at 100 kHz over the same time frame.  All the data is saved in .tdms format, and covers three stator conditions: healthy, inter-turn short circuit (ITSC) fault, and inter-coil short circuit (ICSC) fault.
\subsubsection{Bearing Dataset Description}
The database for bearing faults was collected from vibration sensors mounted on SpectraQuest’s Machinery Fault Simulator (MFS) ABVT system. These time series cover four different simulated states—from normal operation to various fault conditions like healthy, cage, inner, and outer bearing issues. The experimental setup features a 1/4 hp motor that runs between 700 and 3600 rpm.  The bearings are positioned 390 mm apart, and the assembly includes eight balls (each 0.7145 cm in diameter) along with a cage that has a diameter of 2.8519 cm. The data acquisition process is managed by two National Instruments NI 9234 modules. Each module offers four analog acquisition channels and sample data at a rate of 51.2 kHz.

\begin{figure}[htbp]
    \centering
    
    \begin{subfigure}[b]{0.08\textwidth}
        \includegraphics[width=\linewidth,height=1.4cm]{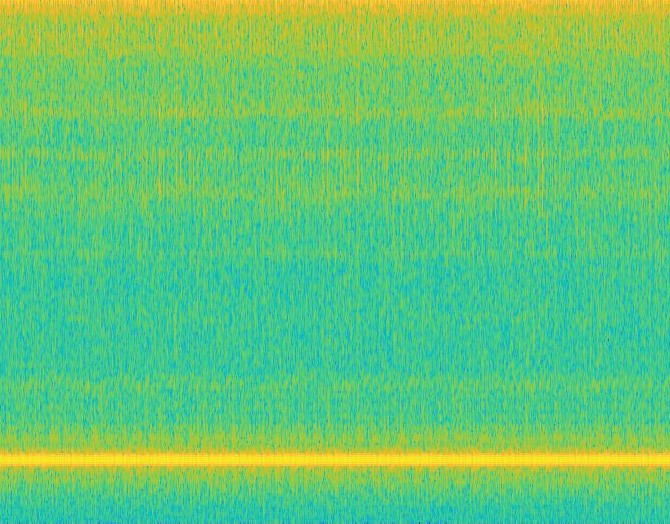}
        \caption{Healthy}
        \label{fig:bearing_healthy}
    \end{subfigure}\hfill
    \begin{subfigure}[b]{0.08\textwidth}
        \includegraphics[width=\linewidth,height=1.4cm]{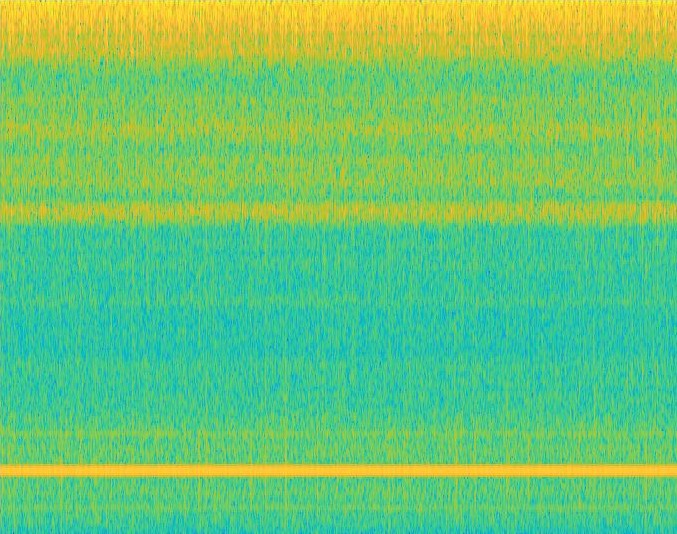}
        \caption{Cage}
        \label{fig:bearing_cage}
    \end{subfigure}\hfill
    \begin{subfigure}[b]{0.08\textwidth}
        \includegraphics[width=\linewidth,height=1.4cm]{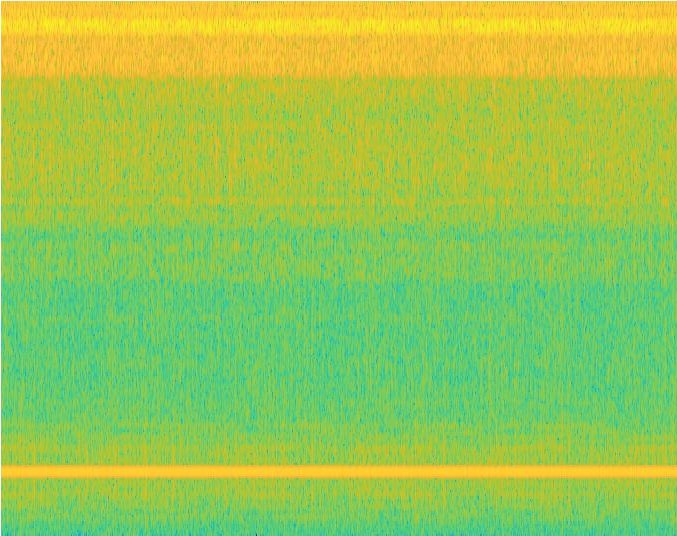}
        \caption{Ball}
        \label{fig:bearing_ball}
    \end{subfigure}\hfill
    \begin{subfigure}[b]{0.08\textwidth}
        \includegraphics[width=\linewidth,height=1.4cm]{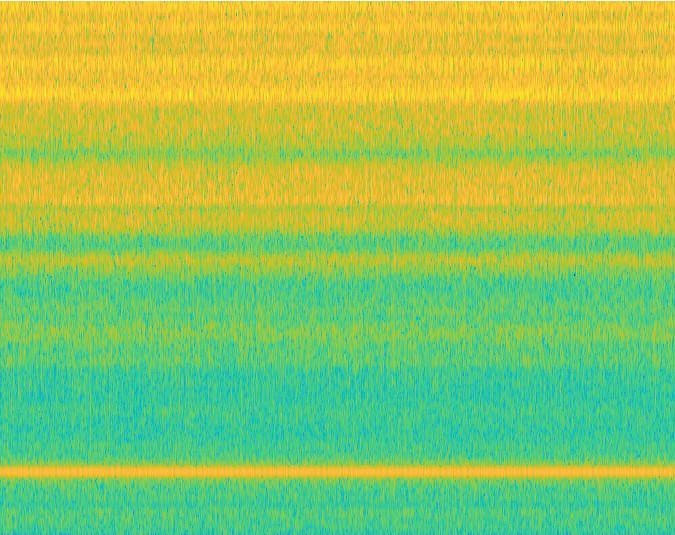}
        \caption{Outer}
        \label{fig:bearing_outer}
    \end{subfigure}
    
    \vspace{0.3em}
    
    \begin{subfigure}[b]{0.08\textwidth}
        \includegraphics[width=\linewidth,height=1.4cm]{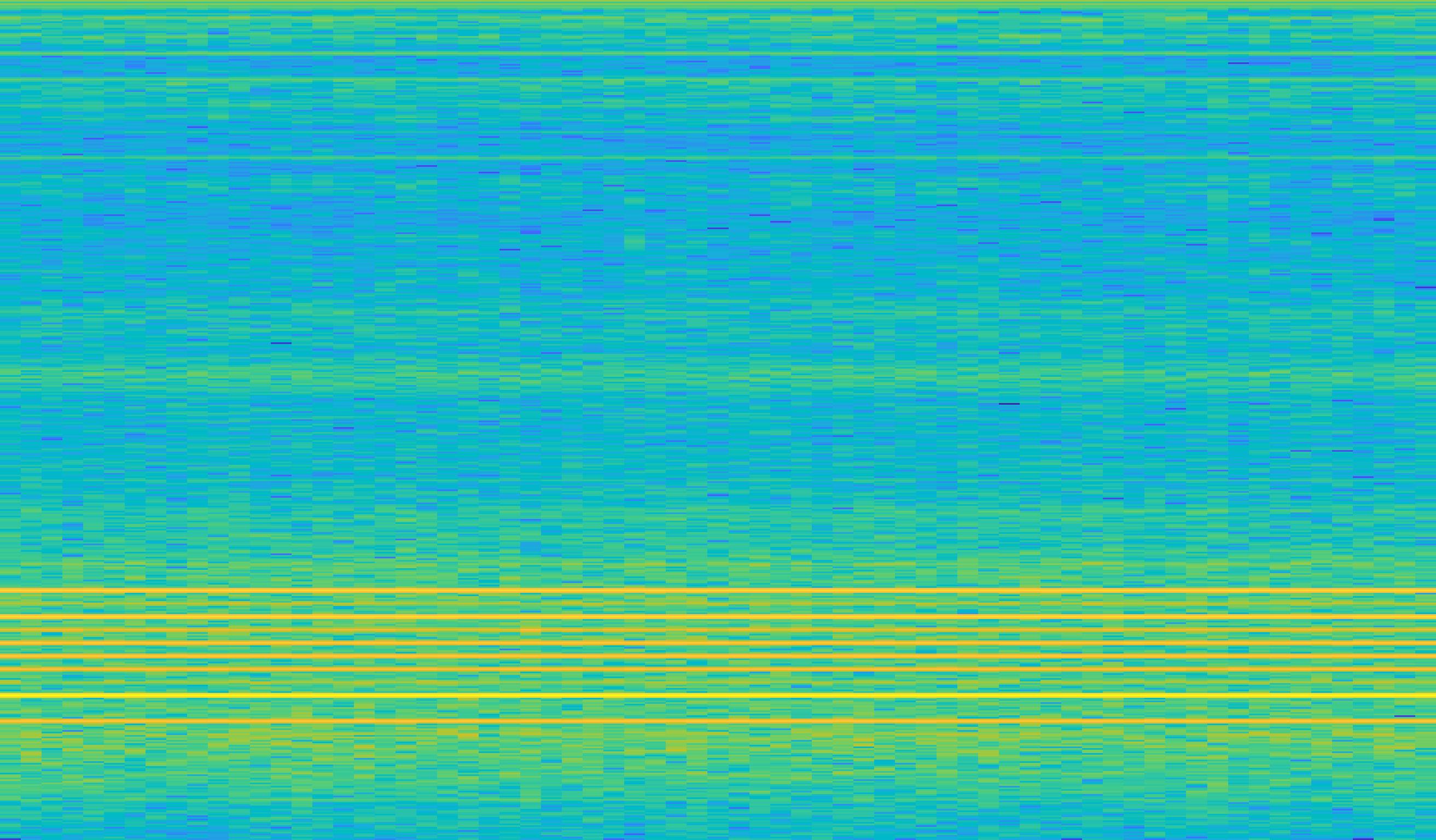}
        \caption{Healthy}
        \label{fig:stator_healthy}
    \end{subfigure}\hfill
    \begin{subfigure}[b]{0.08\textwidth}
        \includegraphics[width=\linewidth,height=1.4cm]{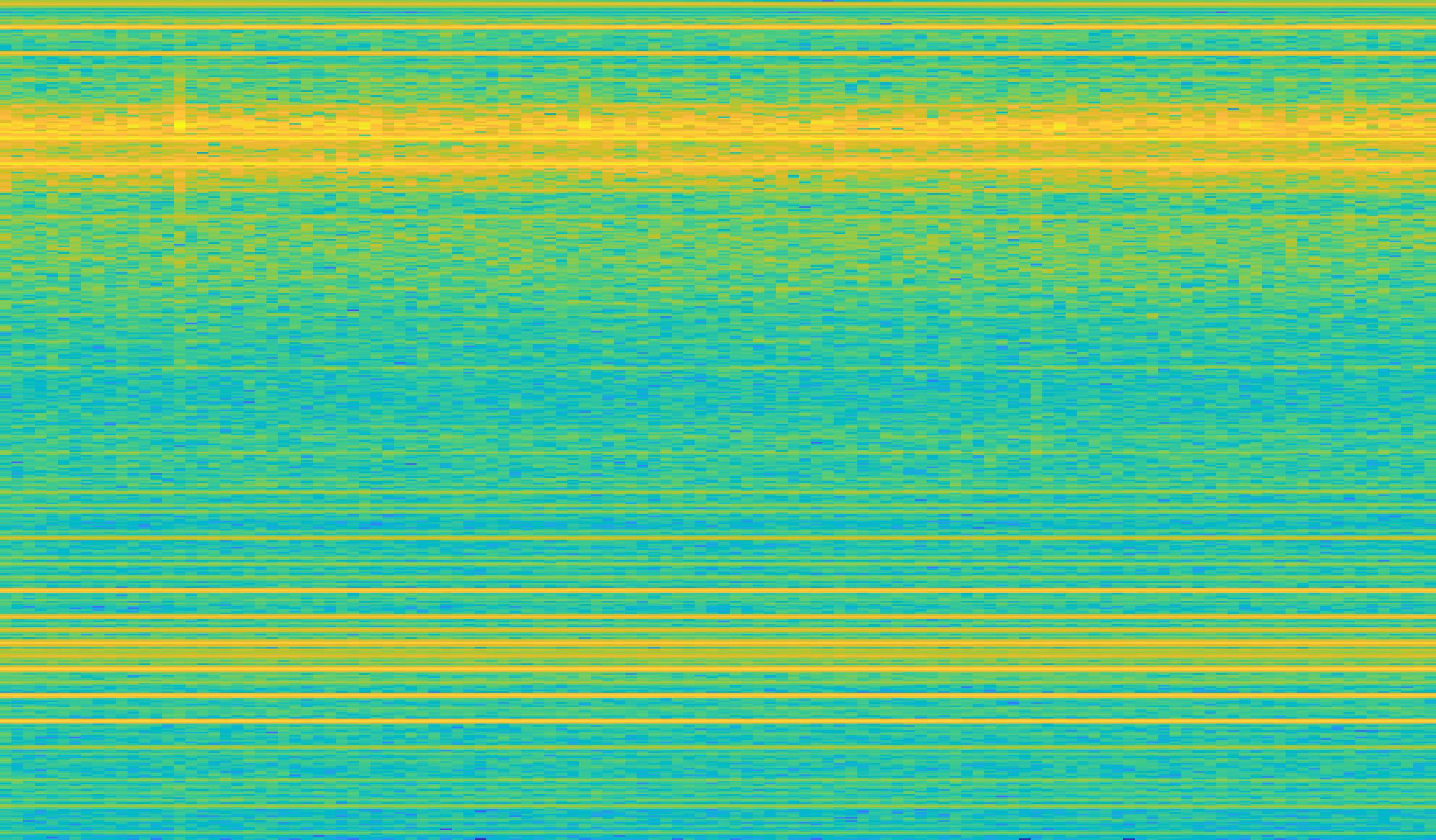}
        \caption{ITSC}
        \label{fig:stator_itsc}
    \end{subfigure}\hfill
    \begin{subfigure}[b]{0.08\textwidth}
        \includegraphics[width=\linewidth,height=1.4cm]{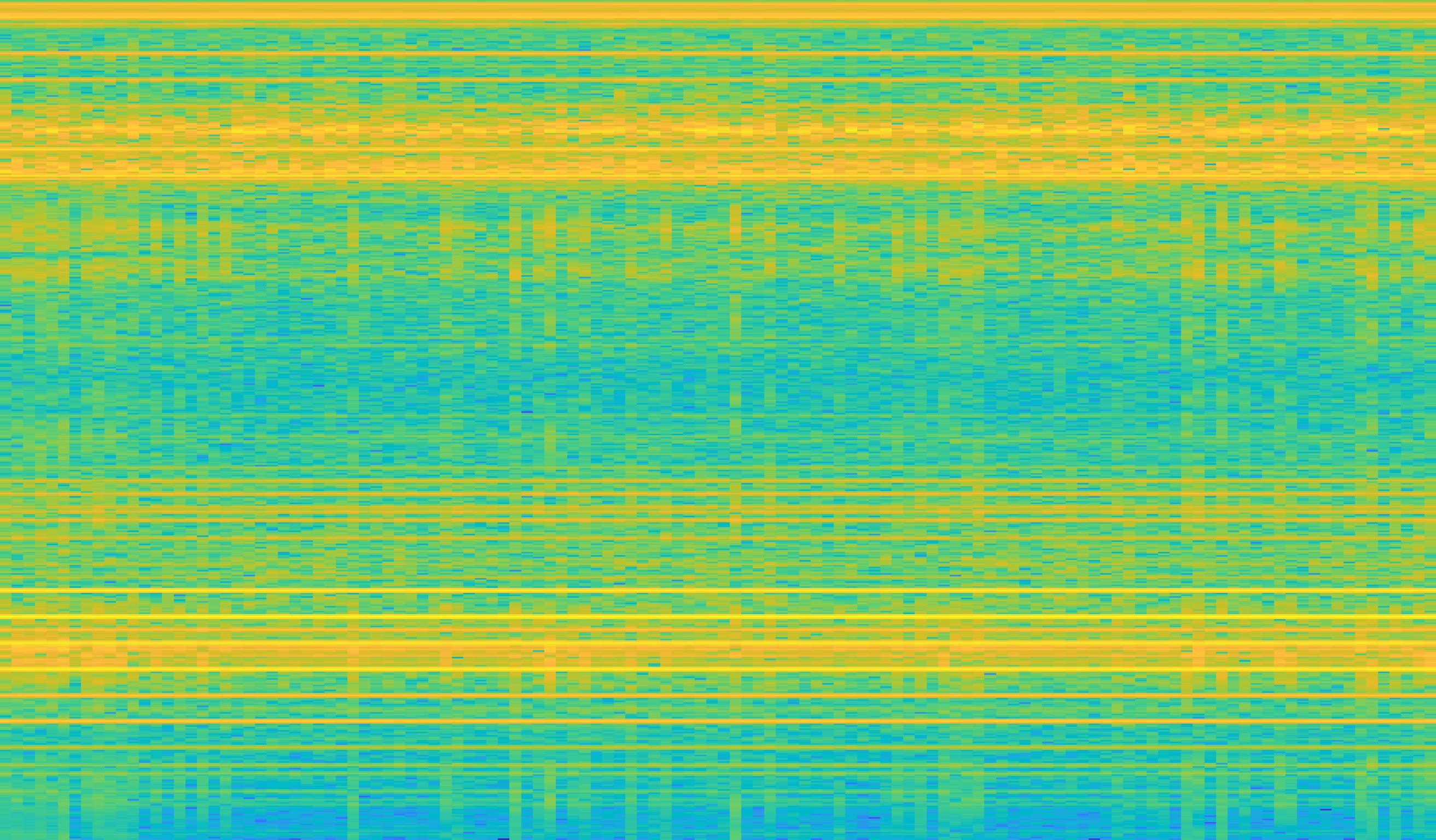}
        \caption{ICSC}
        \label{fig:stator_icsc}
    \end{subfigure}\hfill
    \begin{subfigure}[b]{0.08\textwidth}
        \includegraphics[width=\linewidth,height=1.4cm]{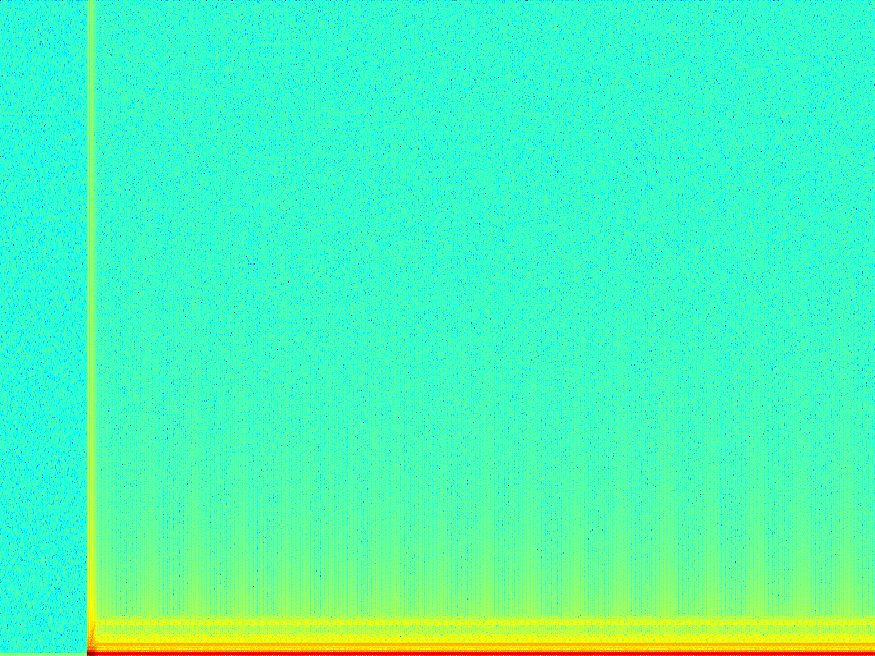}
        \caption{Healthy}
        \label{fig:rotor_healthy}
    \end{subfigure}

    \begin{subfigure}[b]{0.08\textwidth}
        \includegraphics[width=\linewidth,height=1.4cm]{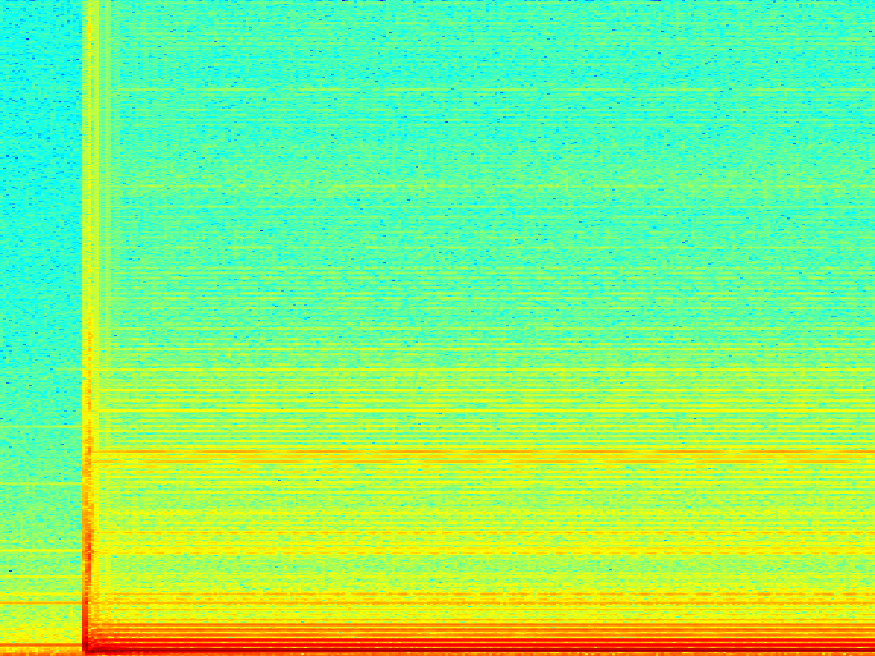}
        \caption{BRB1}
        \label{fig:rotor_brb1}
    \end{subfigure}\hfill
    \begin{subfigure}[b]{0.08\textwidth}
        \includegraphics[width=\linewidth,height=1.4cm]{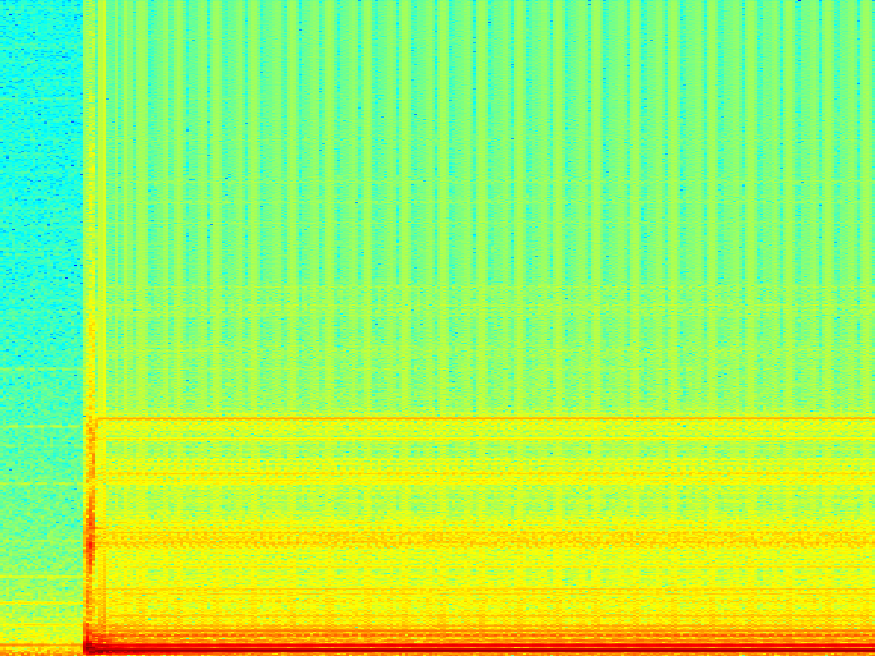}
        \caption{BRB2}
        \label{fig:rotor_brb2}
    \end{subfigure}\hfill
    \begin{subfigure}[b]{0.08\textwidth}
        \includegraphics[width=\linewidth,height=1.4cm]{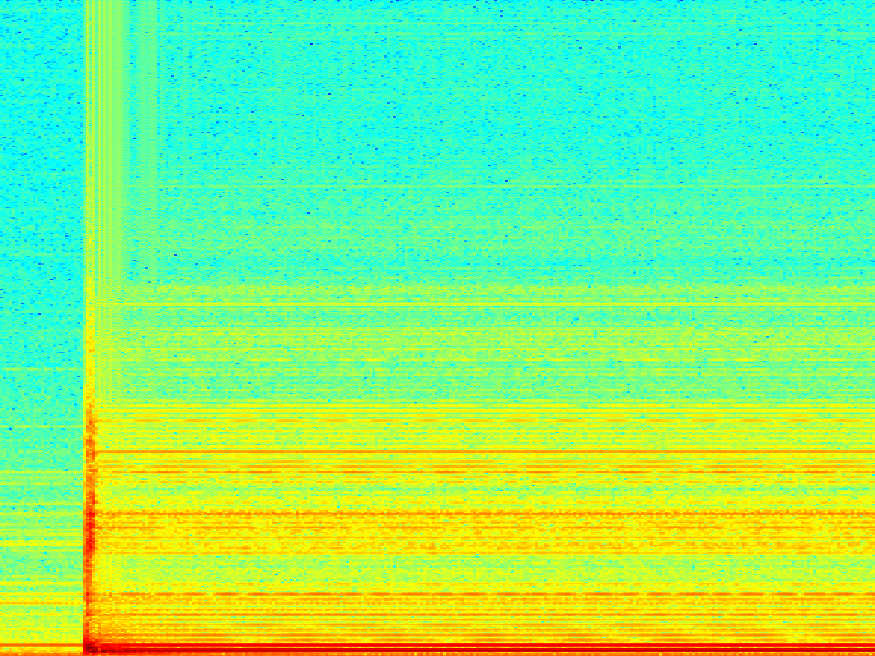}
        \caption{BRB3}
        \label{fig:rotor_brb3}
    \end{subfigure}\hfill
    \begin{subfigure}[b]{0.08\textwidth}
        \includegraphics[width=\linewidth,height=1.4cm]{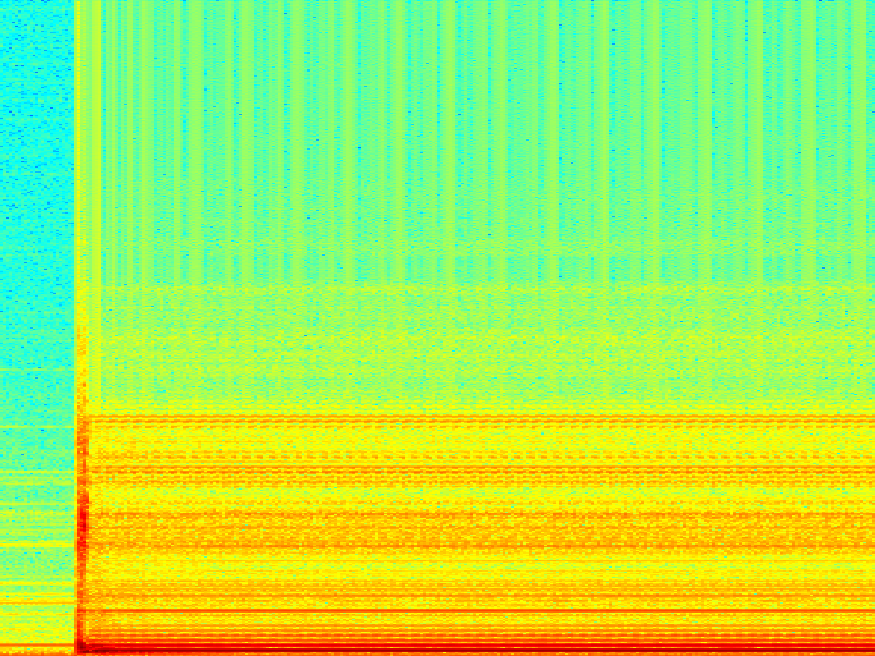}
        \caption{BRB4}
        \label{fig:rotor_brb4}
    \end{subfigure}
    
    \caption{STFT analysis of various motor component conditions: 
    (a-d) Bearing health states, 
    (e-h) Stator fault conditions and rotor health, 
    (i-l) BRB fault scenarios. 
    All plots share identical time-frequency scales.}
    \label{fig:stft_analysis}
\end{figure}

\subsection{Training Details}
The proposed architecture has been evaluated on publicly available multiclass benchmark datasets for industrial machinery condition monitoring \cite{data1,data2,data3}. All experiments have been conducted on a high-performance computing system equipped with an Intel Xeon 3.20 GHz processor (32-core), 128 GB RAM, and an NVIDIA RTX 3080 Ti GPU with 12 GB VRAM. The architecture has demonstrated performance across distinct fault scenarios: bearing, stator, and rotor. 

Hyperparameters are selected through a combination of empirical evaluation and grid search optimisation. The learning rate (0.0001) has been determined by testing values from (0.1, 0.01, 0.001, 0.0005), ensuring optimal convergence without overfitting. The triplet loss margin $ \alpha = 0.27 $ is chosen based on experiments balancing feature separation and training stability. KNN hyperedge formation used $ K = 5 $ after evaluating (3, 5, 7, 10), on similarity threshold at 0.90 by optimizing graph sparsity and connectivity. The model has been trained for 200 epochs, with a batch size of 32, and image input dimensions of 224×224 pixels. These hyperparameter choices ensured stable training while maintaining high classification accuracy.
These settings have been maintained throughout the experiments to ensure uniformity and comparability of results.

\subsection{STFT Hyperparameter Details}
Fo BRB signals in \SI{60}{\hertz} IMs, STFT employs a \SI{200}{\milli\second} window (2000 samples at \SI{10}{\kilo\hertz} sampling), \SI{75}{\percent} overlap (1500 samples), and a 2048-point FFT (\SI{\sim4.88}{\hertz} resolution) within the \SIrange{0}{200}{\hertz} range using a Hann window to minimize leakage and identify rotor asymmetry sidebands.

For bearing dataset, STFT uses a \SI{5}{\milli\second} window (256 samples at \SI{51.2}{\kilo\hertz}), \SI{75}{\percent} overlap (192 samples), and a 512-point FFT (\SI{\sim100}{\hertz} resolution) within the \SIrange{0}{10}{\kilo\hertz} range using a Blackman-Harris window to resolve high-frequency transients under \SI{\pm50}{\gram} accelerometer limits.  

For stator faults, vibration data uses a \SI{20}{\milli\second} window (512 samples at \SI{25.6}{\kilo\hertz}), \SI{70}{\percent} overlap, and a 1024-point FFT (\SI{\sim25}{\hertz} resolution) for \SIrange{0}{2.5}{\kilo\hertz} vibrations, while current data employs a \SI{50}{\milli\second} window (5000 samples at \SI{100}{\kilo\hertz}) and 8192-point FFT (\SI{\sim12.2}{\hertz} resolution) to isolate \SIrange{0}{1}{\kilo\hertz} modulation sidebands, both with Hann windows.  

A total of 50,000 STFT spectral spectrograms have been generated from the temporal IM signals. Each class STFT is used for training and testing the model performance as shown in Figure \ref{fig:stft_analysis}. 


\section{Results and Discussion}\label{sec:res}

\subsection{Performance on Individual Fault Categories}
\begin{figure}[htbp]
    \centering
    \includegraphics[width=0.48\textwidth]{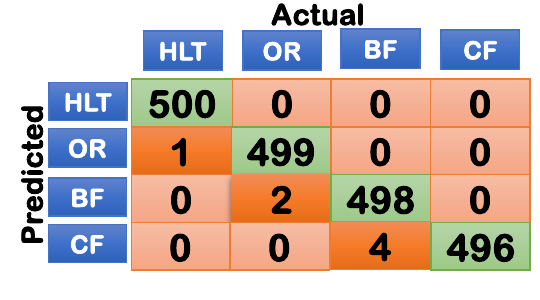} 
    \caption{Confusion Matrix of Bearing Vibration Dataset}
    \label{fig5}
\end{figure}

In the initial experiment, a bearing fault dataset, comprising four distinct operational states, has been employed: Healthy (HLT), Ball Fault (BF), Outer Race Fault (OR), and Cage Fault (CF). The dataset includes 2500 samples for each category (HLT, BF, OR, CF) with a train/test split of 80/20, respectively. Each sample has been represented in two formats: raw time-series signals and STFT spectrograms. Its performance has been evaluated using a confusion matrix (CM), as depicted in Figure \ref{fig5}.

The matrix illustrates the distribution of predicted versus actual values for each class. All test samples of the HLT class have been accurately classified, demonstrating the model's perfect performance in identifying non-faulty conditions. For the OR class, 499 out of 500 samples have been correctly predicted, with only one OR sample misclassified as healthy. Similarly, for the BF class, the model has correctly predicted 498 out of 500 samples, with two instances misclassified, predominantly as OR. In the case of the CF class, 496 out of 500 test samples have been correctly identified, with four samples misclassified as BF. The overall accuracy of the model on the bearing fault dataset is 99.61\%, highlighting its high classification performance.

\begin{figure}[htbp]
  \centering
  
  \begin{subfigure}{0.8\linewidth}
    \centering
    \includegraphics[width=\linewidth]{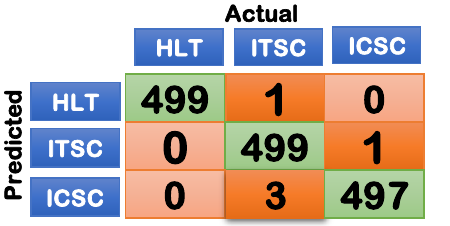}
    \caption{Confusion Matrix of Stator Current Dataset}
    \label{fig6a}
  \end{subfigure}
  
  \vspace{0.5em} 
  
  \begin{subfigure}{0.8\linewidth}
    \centering
    \includegraphics[width=\linewidth]{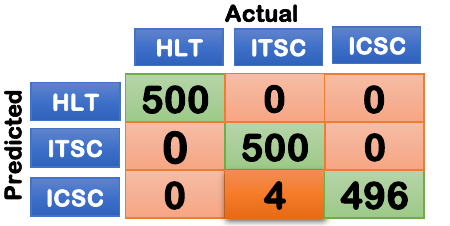}
    \caption{Confusion Matrix of Stator Vibration Dataset}
    \label{fig6b}
  \end{subfigure}
  
  \caption{Confusion matrices: (a) Stator Current Analysis (b) Stator Vibration evaluation using MM-HCAN model.}
  \label{fig:Ensemble_Rotor_comparison}
\end{figure}

For the second experiment, the stator fault dataset, which includes both current and vibration signals, is utilised. This dataset comprises three distinct operational states: Healthy (HLT), inter-turn short circuit (ITSC), and inter-coil short circuit (ICSC). A total of 7500 samples, 6000 samples are used for training and 1500 samples are used for testing on both current and vibration signals separately. The performance of the trained model has been assessed using CM, as illustrated in Figures \ref{fig6a} and \ref{fig6b}. Specifically, Figure \ref{fig6a} presents the results based on current signals, while Figure \ref{fig6b} shows the results based on vibration signals. The model demonstrated exceptional performance when evaluated with current signals, as shown in Figure \ref{fig6a}. It accurately classified 499 out of 500 healthy and ITSC samples, and 497 out of 500 ICSC samples. These results highlight the model's high precision in distinguishing between the operational states based on current signals. Similarly, when evaluated on unseen vibration signals shown in figure \ref{fig6b}, the model accurately classified all healthy and ITSC samples, with minimal misclassifications of ICSC samples as ITSC. The overall accuracy of the model for the stator current and vibration faults diagnosis is   99.61\% and 99.69\%, respectively, reflecting the model's ability to identify faults with high precision.

\begin{figure}[htbp]
  \centering
  
  \begin{subfigure}{0.95\linewidth}
    \centering
    \includegraphics[width=\linewidth]{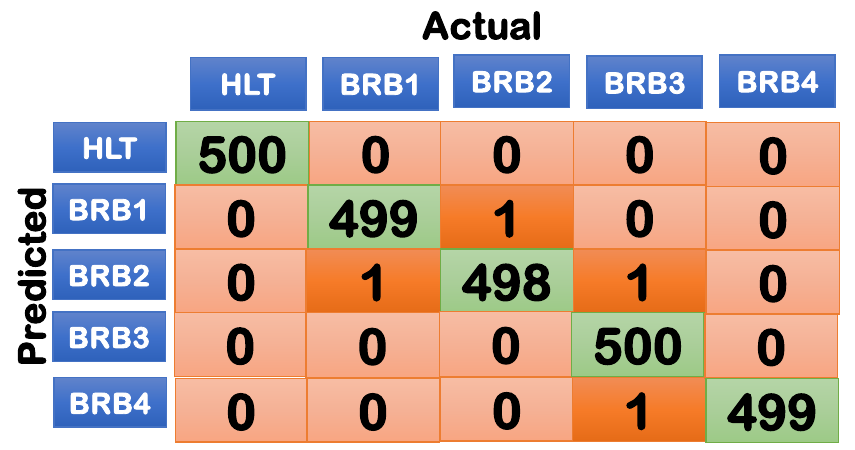}
    \caption{Confusion Matrix of Rotor Current Dataset}
    \label{fig7a}
  \end{subfigure}
  
  \vspace{1em} 
  
  \begin{subfigure}{0.95\linewidth}
    \centering
    \includegraphics[width=\linewidth]{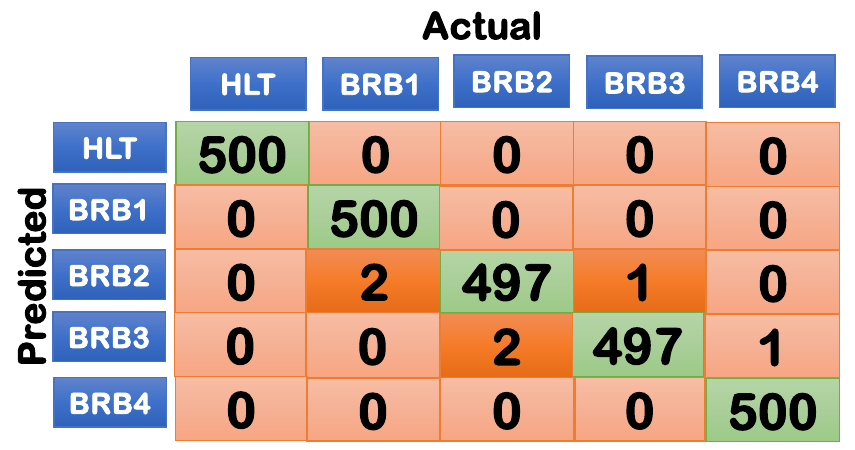}
    \caption{Confusion Matrix of Rotor Vibration Dataset}
    \label{fig7b}
  \end{subfigure}
  
  \caption{Confusion matrices: 
  (a) Rotor Current Analysis, 
  (b) Rotor vibration Analysis using MM-HCAN model.}
  \label{fig:Another_Ensemble_comparison}
\end{figure}

\newcommand{\cmark}{\checkmark}  
\newcommand{\xmark}{\text{\sffamily X}}  
\begin{table*}[t]
\centering
\caption{Comparison of Our Approach (MM-HCAN) with Other Techniques Across Bearing, Stator, and Rotor Datasets}
\label{tab:unified_comparison}
\resizebox{\textwidth}{!}{%
\begin{tabular}{cccccccccccccc}
\toprule
\multirow{2}{*}{\textbf{Ref}} & \multicolumn{4}{c}{\textbf{Bearing Faults}} & \multicolumn{3}{c}{\textbf{Stator Faults}} & \multicolumn{3}{c}{\textbf{Rotor Faults}} & \multirow{2}{*}{\textbf{Measurements}} & \multicolumn{2}{c}{\textbf{Model Accuracy (\%)}} \\ 
\cmidrule(lr){2-5} \cmidrule(lr){6-8} \cmidrule(lr){9-11} \cmidrule(lr){13-14}
 & \textbf{HLT} & \textbf{OR} & \textbf{BR} & \textbf{CF} & \textbf{HLT} & \textbf{ITSC} & \textbf{ICSC} & \textbf{1 BRB} & \textbf{2 BRB} & \textbf{Mult. BRB} &  & \textbf{Current} & \textbf{Vibration} \\ \midrule

\cite{br2} & \cmark & \cmark & \cmark & \cmark & - & - & - & - & - & - & Vibration & - & 98.50 \\ \midrule
\cite{br3} & \cmark & \cmark & \cmark & \cmark & - & - & - & - & - & - & Vibration & - & 99.18 \\ \midrule
\cite{gnn2} & \cmark & \cmark & \cmark & \cmark  &  &  &  & - & - & - & Vibration & - & 99.41 \\ \midrule

\cite{st1} & - & - & - & - & \cmark & \cmark & \cmark & - & - & - & Both & 43.20 & 83.00 \\ \midrule

\cite{st2} & - & - & - & - & \cmark & \cmark & \cmark & - & - & - & Current & 98.20 & - \\ \midrule

\cite{rc2} & - & - & - & - & - & - & - & \cmark & \cmark & \cmark & Vibration & - & 97.67 \\ \midrule
\cite{rc3} & - & - & - & - & - & - & - & \cmark & \cmark & \xmark & Current & 95.80 & - \\ \midrule
\cite{rc4} & - & - & - & - & - & - & - & \cmark & \cmark & \xmark & Current & 99.10 &  \\ \midrule
\cite{wpedl} & \cmark & \cmark & \cmark & \cmark & \cmark & \cmark & \cmark & \cmark & \cmark & \cmark & Both & 98.89 & 98.45 \\ \midrule
\textbf{MM-HCAN} & \cmark & \cmark & \cmark & \cmark & \cmark & \cmark & \cmark & \cmark & \cmark & \cmark & \textbf{Both} & \textbf{99.60} & \textbf{99.52} \\ \bottomrule
\end{tabular}}
\end{table*}

The third experiment is executed utilising the IM  current and vibration rotor dataset, which encompasses five distinct operational conditions: Healthy (HLT), and one, two, three, and four broken rotor bars (BRB1, BRB2, BRB3, and BRB4), respectively. The dataset includes 2500 samples for each category (HLT, BRB1, BRB2, BRB3, and BRB4) with a train/test split of 80/20, respectively. The dataset's performance has been evaluated employing a model that had undergone optimal fine-tuning. The outcomes are graphically represented in Figures \ref{fig7a} and \ref{fig7b}. Figure \ref{fig7a} presents the CM associated with the current signals. Analysis of this matrix reveals that the model achieved perfect classification for both the healthy and BRB3 classes, correctly identifying all samples within these categories. The remaining classes—BRB1 and BRB4—exhibited minimal misclassification, with each having only a single instance of incorrect classification. Notably, the BRB2 class displayed two instances of misclassification. Figure \ref{fig7b} shows the test results of BRB on vibration signals. The model perfectly classified all 500 samples of the HLT class. For the fault categories, the model's performance is as follows: 500 correct identifications for BRB1, 497 for BRB2, 497 for BRB3, and all 500 samples correctly identified for BRB4. This indicates a robust performance with minimal misclassifications, primarily between BRB2 and BRB.

\subsection{Cross-Domain Generalisation Performance}
\begin{figure}[htbp]
    \centering
    \includegraphics[width=0.5\textwidth]{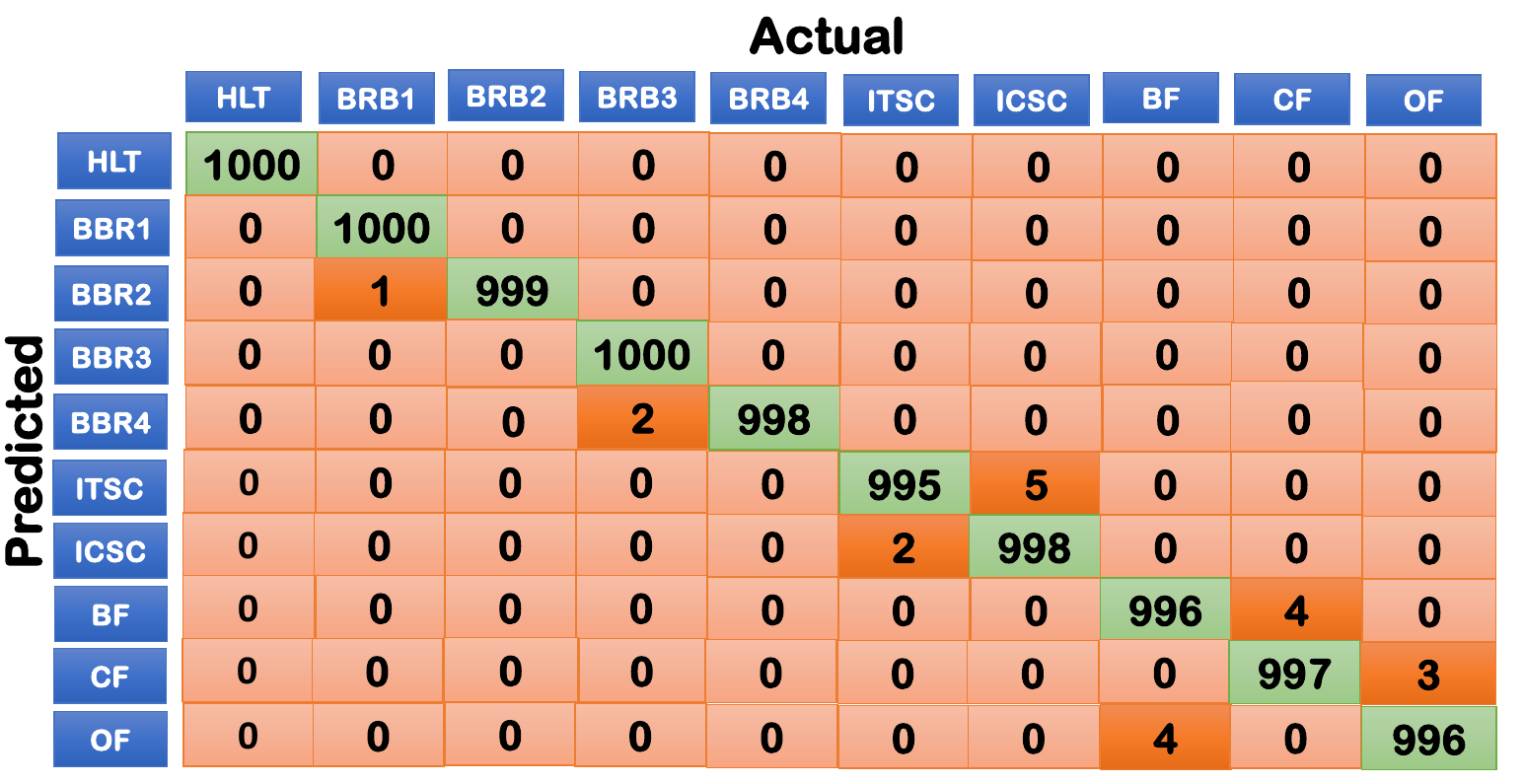}  
    \caption{Confusion matrix analysis of the combined dataset showing classification performance across all operational conditions. The diagonal elements represent correct predictions while off-diagonal values indicate misclassifications.}
    \label{fig-combined}
\end{figure}

In this experimental phase, our proposed model's capacity for generalisation has been assessed on a cross-domain combined fault dataset. This dataset amalgamates various fault types, with 80\% of the data (comprising 40,000 signals) allocated for training purposes, while the remaining 20\% (10,000 signals) are reserved for evaluating the model's performance on previously unseen data. The CM illustrated in Figure \ref{fig-combined} presents the model's generalisation outcomes on the integrated dataset, offering insights into its ability to accurately classify faults across different domains. The model exhibited a high degree of accuracy, correctly identifying all 1000 instances of the HLT class and achieving perfect classification rates for both BRB1 and BRB3, with 1000 accurate predictions for each. The BRB2 class is nearly perfectly classified, with a minor exception of one misclassification as BRB4, while BRB4 had a marginally higher error rate with two misclassifications as BRB3. The ITSC class demonstrated a high level of accuracy, with 995 correct predictions and five instances misclassified as ICSC. The ICSC class has been classified with perfect accuracy, indicating the model's robustness in identifying this particular fault condition. The BF, OF, and CF classes also showed commendable performance, with 996, 996, and 997 correct predictions, respectively, and a minimal number of misclassifications between these classes. These findings highlight the model's overall efficacy in distinguishing between diverse motor conditions.

\subsection{Comparative Analysis}
In Table \ref{tab:unified_comparison}, we present a comprehensive comparison of MM-HCAN with existing State-of-the-Art (SOTA) techniques. The initial four investigations were executed on a bearing vibration dataset, where the highest recorded accuracy was 99.41\%. Subsequent analyses were carried out on stator current and vibration datasets, yielding maximum accuracies of 98.20\% for current signals and 83\% for vibration signals. Further evaluations were performed on rotor datasets, with the peak accuracies reaching 99.10\% for current data and 98.45\% for vibration signals. Our model demonstrates superior performance overall compared to SOTA techniques, achieving an accuracy of 99.60\% on current signals and 99.52\% on vibration signals, thereby establishing a new benchmark in this domain.

\subsection{Ablation Study}
The ablation study detailed in Table \ref{tab5} presents a thorough examination of the performance metrics of the MM-HCAN model across various configurations, highlighting the impact of different architectural blocks on classification outcomes. The study systematically varies the inclusion of temporal features ($w_{t}$), spectral features ($w_{s}$), cross-domain features ($w_{cr}$), contrastive loss ($w_{cl}$), absence of contrastive loss ($w_{ncl}$), and the attention mechanism ($w_{att}$) to evaluate their individual contributions. The performance metrics, including accuracy, precision, recall, F1-score, and area under the curve (AUC), are reported for each configuration.
\begin{table}[htbp] 
\centering
\caption{Performance Metrics of MM-HCAN Across Various Ablation Experiments} 
\label{tab5}
\resizebox{\columnwidth}{!}{
{
    \tiny 
    \renewcommand{\arraystretch}{0.8} 
    \setlength{\tabcolsep}{2pt}      

    \begin{tabular}{@{}*{6}{c} *{5}{c}@{}} 
        \toprule
        \multicolumn{6}{c}{\textbf{Architecture Blocks}} & \multicolumn{5}{c}{\textbf{Classification Metrics}} \\
        \cmidrule(r){1-6} \cmidrule(l){7-11}
        \textit{$w_t$} & \textit{$w_s$} & \textit{$w_{cr}$} & \textit{$w_{cl}$} & \textit{$w_{ncl}$} & \textit{$w_{att}$} & \textbf{Acc} & \textbf{Pre} & \textbf{Rec} & \textbf{F1} & \textbf{AUC} \\
        \midrule
        $\checkmark$ & & & & $\checkmark$ &  & 91.87 & 93.05 & 91.87 & 91.68 & 94.20 \\
        & $\checkmark$ & & & $\checkmark$ & & 94.23 & 95.30 & 94.27 & 94.19 & 96.22 \\
        & & $\checkmark$ & & $\checkmark$ & & 96.11 & 96.95 & 96.17 & 96.61 & 96.62 \\
        $\checkmark$ & $\checkmark$ & & & & & 97.11 & 97.92 & 97.51 & 97.11 & 97.31 \\
        $\checkmark$ & $\checkmark$ & & $\checkmark$ & & & 97.19 & 97.24 & 97.15 & 97.19 & 97.18 \\
        $\checkmark$ & $\checkmark$ & $\checkmark$ & & & & 97.85 & 97.43 & 97.51 & 97.11 & 97.90 \\
        $\checkmark$ & $\checkmark$ & $\checkmark$ &$\checkmark$ &  & & 98.22 & 98.69 & 98.72 & 98.81 & 98.21 \\
        $\checkmark$ & $\checkmark$ & $\checkmark$ & $\checkmark$ &  & $\checkmark$ & \textbf{99.47} & \textbf{99.25} & \textbf{99.28} & \textbf{99.33} & \textbf{99.49} \\
        \bottomrule
    \end{tabular}%
}
}
\end{table}

The baseline model, incorporating only temporal features, achieves an accuracy of 91.87\%, with a corresponding precision and recall of 93.05\% and 91.87\%, respectively. The inclusion of spectral features increases the model's accuracy to 94.23\%, indicating the complementary nature of these features in enhancing predictive performance. The integration of cross-domain features further refines the model, achieving an accuracy of 96.11\%, underlining the importance of domain-invariant learning for robust classification. The addition of the contrastive loss mechanism with the attention mechanism yields incremental improvements, with the fully integrated model (including all features and mechanisms) attaining the highest accuracy of 99.47\%. This configuration also demonstrates superior precision (99.25\%), recall (99.28\%), F1-score (99.33\%), and AUC (99.49\%), showcasing the synergistic effect of combining multiple architectural elements. The findings from this ablation study underscore the critical role of each architectural component in optimising the MM-HCAN model's performance. The attention mechanism, in particular, emerges as a pivotal feature, significantly enhancing the model's capacity to discern subtle distinctions between classes. These results validate that each component, rather than being arbitrarily combined, adds distinct and incremental diagnostic value, supporting a systematic design rationale.

\subsection{Comparative Analysis with Hybrid Architectures}

The domain of fault diagnosis in IMs has seen diverse hybrid approaches (Section \ref{sec:introduction}, Table \ref{tab:unified_comparison}), and our ablation study (Table \ref{tab5}) confirms the individual contribution of each constituent block within MM-HCAN. We also conducted a further benchmark to illustrate MM-HCAN's distinct advantages. For this comparative analysis, the baseline hybrid models (CNN+LSTM, GCN+LSTM, CNN+GCN, CNN+LSTM+GCN) have been implemented using representative architectures for each module to ensure a rigorous evaluation. Specifically, the CNN components in these baselines utilised a VGG16 architecture. The GCN components comprised a (3-layer GCN with 128 hidden units per layer and ReLU activation), and the LSTM with 128 hidden units. These baselines are then enhanced with contrastive learning or attention mechanisms as specified in Table \ref{tab3:hybrid_comp}.
Despite their architectural depth and the inclusion of these advanced components, the established hybrid strategies (the strongest performing alternative, CNN+LSTM+GCN with attention and contrastive learning, yielded 97.88\% Acc, and 97.91\% F1) do not reach the performance levels of MM-HCAN (99.47\% Acc, and 99.49\% F1).  MM-HCAN’s architecture, driven by unique contributions, delivers a marked improvement in efficacy, especially for challenging cross-domain generalisation tasks, indicating a clear advancement over aggregated hybrid approaches.

\begin{table}[h]
\centering
\caption{Hybrid Architectures Vs MM-HCAN}
\label{tab3:hybrid_comp}
\resizebox{\columnwidth}{!}{%
    \scriptsize
    \renewcommand{\arraystretch}{0.85}
    \setlength{\tabcolsep}{3pt}
   
    \begin{tabular}{@{} : c : ccccccc : cc @{}}
        \toprule
        
        & \multicolumn{7}{c}{\textbf{Architecture Blocks}} & \multicolumn{2}{c}{\textbf{Metrics}} \\
        
        \cmidrule(lr){2-8} \cmidrule(l){9-10}
        \textbf{Hybrid Architectures} & \textit{$w_t$} & \textit{$w_s$} & \textit{$w_{cr}$} & \textit{$w_g$} & \textit{$w_{cl}$} & \textit{$w_{ncl}$} & \textit{$w_{att}$} & \textbf{Acc}  & \textbf{F1}  \\
        \midrule
        
        \multirow{4}{*}{\textbf{CNN + LSTM}}
        & $\checkmark$ & & & & & $\checkmark$ &  & 75.20  & 74.32  \\
        & & $\checkmark$ & & & & $\checkmark$ &  & 87.5 & 87.10 \\
        & $\checkmark$ & $\checkmark$ & $\checkmark$ & &  & & $\checkmark$ & 92.31 & 92.27  \\
        & $\checkmark$ & $\checkmark$ & $\checkmark$ & & $\checkmark$ & &  & 94.44  & 93.89 \\
        & $\checkmark$ & $\checkmark$ & $\checkmark$ & & $\checkmark$ & & $\checkmark$ & 95.66  & 95.61 \\
        
        \midrule
        
        \multirow{5}{*}{\textbf{GCN + LSTM}}
        &  &  & & $\checkmark$ & & $\checkmark$ &  & 82.40  & 81.31 \\
        &  $\checkmark$ &  &  & $\checkmark$ & & $\checkmark$ &  & 90.30 & 91.47 \\
        & $\checkmark$ &  &  & $\checkmark$ & $\checkmark$ & &  & 95.10 & 94.36 \\
        & $\checkmark$ &  & $\checkmark$ & $\checkmark$ & $\checkmark$ &  & & 95.40 & 95.56\\
        & $\checkmark$ &  & $\checkmark$ & $\checkmark$ & $\checkmark$ &  & $\checkmark$ & 96.59 & 96.11 \\
        \midrule
       
        \multirow{5}{*}{\textbf{CNN + GCN}}
        &  & $\checkmark$ &  & $\checkmark$ &  & $\checkmark$ &  & 93.72 & 93.11 \\ 
        &  & $\checkmark$ & $\checkmark$  & $\checkmark$ &  & $\checkmark$ &  & 96.51 & 96.07 \\
        &  & $\checkmark$ & $\checkmark$  & $\checkmark$ & $\checkmark$ &  &  &  97.15 & 96.98  \\
        &  & $\checkmark$ & $\checkmark$  & $\checkmark$ & $\checkmark$ &  & $\checkmark$ &  97.22& 97.13 \\
        \midrule
       
        \multirow{4}{*}{\textbf{CNN + LSTM + GCN}}
        & $\checkmark$ & $\checkmark$ &  $\checkmark$ & $\checkmark$ &  & $\checkmark$ &  &  97.45 & 97.22 \\ 
        &  $\checkmark$ & $\checkmark$ & $\checkmark$  & $\checkmark$ &  & $\checkmark$ & $\checkmark$ & 97.68 & 97.43 \\
        & $\checkmark$ & $\checkmark$ & $\checkmark$  & $\checkmark$ & $\checkmark$ &  &  & 97.81 & 97.97 \\
        & $\checkmark$ & $\checkmark$ & $\checkmark$  & $\checkmark$ & $\checkmark$ &  & $\checkmark$ & 97.88 & 97.91 \\
        \midrule
        
        \multirow{1}{*}{\textbf{MM-HCAN}}
        & $\checkmark$ & $\checkmark$ & $\checkmark$ &  & $\checkmark$ &  & $\checkmark$ & \textbf{99.47} & \textbf{99.49} \\ 
        \bottomrule
    \end{tabular}%
}
\end{table}

\subsection{Robustness Test}

In real-world industrial environments, sensor data is often subject to noise and external disturbances on current and vibration signals. To ensure our MM-HCAN model can handle these challenges, we tested its performance under three common types of noise. First, we added Gaussian noise ($\sigma = 0.224$, SNR=10 dB ) to the signals. The model’s accuracy dropped slightly, from 99.60\% (on clean data) to 98.86\% in cross-domain classification tasks. Next, we introduced harmonic distortion by extra frequency components ( 3\textsuperscript{rd}, 5\textsuperscript{th}, and 7\textsuperscript{th} harmonics, each with 20\% amplitude) caused by non-linear loads, like those from variable-speed drives. Here, accuracy stayed high at 98.28\%, proving the model can handle distortions from power system irregularities. Finally, we tested sudden spikes (20\% of a normalised signal’s maximum amplitude) in both IMs signals. Even in this harsh scenario, accuracy remained robust at 98.05\%, highlighting the system’s ability to ignore short-lived disruptions. Overall, the total decline in accuracy across all tests is less than 1.5\%. Misclassifications have been observed mainly in fault categories with subtle differences between BRB3 vs. BRB4, ITSC vs. ICSC, and CF vs. OF. These results indicate that MM-HCAN maintains reliable performance (above 98\% accuracy) even in noisy industrial environments, making it a practical tool for motor diagnostics.

\subsection{Computational Performance}
To assess real-time deployment feasibility, we evaluated MM-HCAN's computational efficiency. The model processed 40,000 training samples in approximately 5 hours, achieving an inference speed of \SI{5.7}{\milli\second} per sample. Compared to conventional CNN architectures (i.e., VGG16: \SI{7.2}{\milli\second}, ResNet152: \SI{8.4}{\milli\second}, DenseNet264: \SI{12.2}{\milli\second}), MM-HCAN’s hypergraph-driven feature propagation reduces computational overhead by \SIrange{18}{40}{\percent}, delivering faster inference while retaining superior classification accuracy.

\section{Discussion}
The experimental results demonstrate that MM-HCAN consistently outperforms state-of-the-art models across individual and cross-domain fault diagnosis tasks. The integration of hypergraph neural networks with contrastive learning and multi-head attention allows to capture both global and localised relationships within and across modalities, which are critical for accurate fault classification. Particularly notable is MM-HCAN’s ability to maintain high performance in cross-domain generalisation tasks and under noisy signal conditions, which are often challenging for traditional CNN- or LSTM-based architectures.  
The results highlight, combining STFT-based spectral features with temporal patterns captured by 1D CNNs and LSTMs. This is orchestrated by several core innovations unique to MM-HCAN: (1) a structured framework for multimodal fusion leveraging dynamically constructed hyperedges, which allows for more expressive higher-order relationships between modalities; (2) a novel embedding update mechanism via modality-specific hypergraph Laplacians, enabling fine-grained, context-aware feature refinement; and (3) the application of triplet-based contrastive learning directly within a hypergraph topology, rather than conventional Euclidean space, promoting more discriminative representations that respect the complex relational data structure.

\section{Conclusion}\label{sec:con}

This research article introduced the MM-HCAN technique as a robust approach for early-stage fault diagnosis in IMs. By leveraging high-dimensional data extracted from vibration and current features, MM-HCAN demonstrates superior efficacy in diagnosing various fault types encountered in IMs, including bearing, rotor, and stator faults. A comparison with conventional models highlights MM-HCAN's superior performance in fault diagnosis. Furthermore, MM-HCAN achieves high accuracies across different fault types, with accuracies of 99.61\% for bearing faults, 99.82\% and 99.76\% for rotor current and vibration datasets, and 99.61\% and 99.69\% for stator current and vibration datasets, respectively. Evaluation of MM-HCAN's robustness through tests on a combined dataset, which correctly classified 99.47\% of test cases, further solidifies its utility in industrial settings. These findings suggest that MM-HCAN holds significant promise for enhancing industrial operational efficiency and reliability by facilitating early fault detection in IMs.

Future research will focus on several promising directions. 
Architecturally, we plan to explore dynamic hypergraph construction in self-supervised contrastive manners to adapt to evolving fault characteristics and investigate further reduce reliance on labelled datasets. Although attention mechanisms provide some insight into feature relevance, future work could integrate explainable AI (XAI) frameworks to better trace classification decisions, which is particularly valuable for maintenance engineers in high-stakes industrial environments. Extending MM-HCAN to incorporate additional modalities, such as thermal or acoustic signals, could also enhance diagnostic precision for a wider range of incipient faults. From an application perspective, future work includes adapting MM-HCAN for fault severity assessment and remaining useful life (RUL) prediction, providing more comprehensive prognostic capabilities. Furthermore, deploying and validating MM-HCAN in real-time industrial environments on diverse machinery beyond IMs represents a key objective to demonstrate its broader applicability and scalability for industrial adoption.

\bibliography{ref}
\bibliographystyle{IEEEtran}

\end{document}